\documentclass[11pt,a4paper]{article}

\pdfoutput=1
\usepackage[final]{acl}

\usepackage{times,latexsym}
\usepackage{url}
\usepackage{booktabs}
\usepackage{amsmath}
\usepackage{graphicx}
\usepackage{subfig}
\usepackage[T1]{fontenc}

\usepackage[ruled,vlined]{algorithm2e}
\usepackage{listings}
\usepackage{graphicx}
\usepackage{mathtools}
\usepackage{tabularx}
\usepackage{amsmath}
\usepackage{amssymb}
\usepackage{booktabs}
\usepackage{hyperref}
\usepackage[boldmath,np]{numprint}
\usepackage{cleveref}
\usepackage{tablefootnote}
\usepackage{tabularray}
\usepackage{array}

\crefformat{section}{\S#2#1#3}
\crefformat{subsection}{\S#2#1#3}
\crefformat{subsubsection}{\S#2#1#3}
\crefrangeformat{section}{\S\S#3#1#4 to~#5#2#6}
\crefmultiformat{section}{\S\S#2#1#3}{ and~#2#1#3}{, #2#1#3}{ and~#2#1#3}

\title{Context-aware Visual Storytelling with \\ Visual Prefix Tuning and Contrastive Learning}

\author{
  Yingjin Song,  Denis Paperno and Albert Gatt
  \\
  \ \\
 Utrecht University, Utrecht, The Netherlands
  \\
  \texttt{\{y.song5, d.paperno, a.gatt\}@uu.nl}
}

\date{}

\begin{document}
\maketitle
\begin{abstract}
  Visual storytelling systems generate multi-sentence stories from image sequences. In this task, capturing contextual information and bridging visual variation bring additional challenges. We propose a simple yet effective framework that leverages the generalization capabilities of pretrained foundation models, only training a lightweight vision-language mapping network to connect modalities, while incorporating context to enhance coherence. We introduce a multimodal contrastive objective that also improves visual relevance and story informativeness. Extensive experimental results, across both automatic metrics and human evaluations, demonstrate that the stories generated by our framework are diverse, coherent, informative, and interesting.
\end{abstract}

\section{Introduction}\label{sec:intro}



Visual storytelling \cite[VIST;][]{huang_visual_2016} aims at crafting a narrative from a sequence of ordered images.
This task involves a number of key challenges, some of which are well-studied problems in computational narrative generation, while others arise from the visually grounded nature of the task: VIST image sequences exhibit semantic and temporal gaps, so that (i) a successful VIST system needs to balance textual \textbf{coherence} \citep{redeker2000coherence, DBLP:conf/ijcai/CallawayL01} with (ii) visual \textbf{grounding} \citep{wang-etal-2022-rovist, surikuchi-etal-2023-groovist}. At the same time, (iii) generated narratives should capture the reader's attention, necessitating a degree of creativity and \textbf{interestingness} \cite{DBLP:journals/aim/Gervas09}, but should also (iv) be \textbf{informative} \citep{li_informative_2019, chen_commonsense_2021}, that is, incorporate relevant details of the entities and activities in the visual content.


Existing models usually include a vision encoder and language decoder 
either trained from scratch or finetuned \citep{taehyeong_kim_glac_2018, wang_no_2018, hu_what_2020,li_knowledge-enriched_2022, fan_visual_2022, yang_attractive_2023, wang-etal-2024-sco} on the VIST task.
This requires a large amount of computational resources. 
Instead, we propose to benefit
from pre-trained models that have already learned meaningful representations from vast amounts of data, 
following the ClipCap approach \citep{mokady2021clipcap} that
integrates pretrained CLIP \citep{radford2021learning} and GPT2 \citep{Radford2019LanguageMA} 
via a lightweight mapping network. 
ClipCap trains only the mapping network to construct soft visual prefixes from CLIP embeddings to guide GPT2 to generate text, while both CLIP and GPT2 can be kept frozen.
Although visual prefix tuning has been widely used for image captioning, 
it has not been adapted for visual storytelling, and its potential here is yet to be explored. 

Our new framework  incorporates a context-aware mappping network, while addressing coherence by incorporating previous story sentences.
To enhance visual grounding and informativeness, we employ a multimodal training objective.
We further compare four common decoding strategies (beam, top-$k$, nucleus and contrastive search), 
showing that they have substantial impact on the generation quality, especially as reflected in  human evaluation, in contrast to standard metrics.


The main contributions of this work are:\footnote{Our code and model are available at https://github.com/yjsong22/ContextualVIST}\\
\begin{itemize}   
    \item a framework to incorporate textual coherence in VIST, while leveraging
    pretrained models;
    \item contrastive training 
    to improve
    informativeness and visual grounding;
    \item a comprehensive human evaluation targeting the four challenges outlined above;
    \item extensive evaluation demonstrating competitiveness with state-of-the-art baselines.
\end{itemize}

\section{Related Work}


\paragraph{Visual Storytelling.}
The Visual Storytelling (VIST) task \citep{huang_visual_2016} aims to create narrative continuity between images for a fluent, coherent story. Early attempts extended image captioning models by combining global-local visual attention \citep{taehyeong_kim_glac_2018} and learning contextualized image representations \citep{diana_gonzalez-rico_contextualize_2018}. Considerable efforts explored Reinforcement Learning (RL) with custom reward functions for visual storytelling \citep{jing_wang_show_2018, wang_no_2018, huang_hierarchically_2019, hu_what_2020}.
Given that storytelling involves imagination and reasoning, many works \citep{ yang_knowledgeable_2020, hsu_knowledge-enriched_2020, wang_storytelling_2020, chen_commonsense_2021, xu_imagine_2021, zheng_two_2021, li_knowledge-enriched_2022, wang-etal-2024-sco} also integrate external knowledge to introduce commonsense concepts not directly present in visual input.

Recent research leverages Transformer-based architectures to learn multimodal feature embeddings, integrating image regions with semantic relationships
\citep{qi_latent_2021}.
Several studies have focused on utilizing pre-trained models for visual storytelling, either by fine-tuning pre-trained Transformer encoders \citep{fan_visual_2022}, or jointly tuning pre-trained LMs with pre-trained image encoders \citep{yu_transitional_2021}.
Other variants consider additional factors such as emotion/sentiment \citep{li2019emotion}, personas \citep{chandu-etal-2019-way, DBLP:conf/aaai/LiuK23, 10.1162/tacl_a_00553}, and writing style \citep{DBLP:conf/mm/WangZLL23, yang_attractive_2023}. 
Unlike prior work, our approach efficiently adapts frozen VLMs and LLMs, conditioning on both textual context and visual input to ensure story continuity and coherence.

\paragraph{Prompt and Prefix Tuning.}

Prompting means designing  “instructions” for pretrained 
language models (LM) 
to generate desired outputs, conditioning them on either human-crafted templates or automatically optimized tokens \citep{DBLP:journals/csur/LiuYFJHN23}. 
Much research proposes to automate prompt engineering by learning discrete \citep{10.1162/tacl_a_00324,  haviv-etal-2021-bertese,  10.1162/tacl_a_00468} or continuous prompts \citep{li-liang-2021-prefix, lester_power_2021}. 
The latter 
can be updated via back-propagation, making them less constrained than 
\citep{zhong-etal-2021-factual, petrov2023prompting}.
With large frozen LMs, Prompt Tuning \citep{lester_power_2021} simply adds a tunable, 
real-valued embedding to the input of the decoder, achieving results comparable to full model fine-tuning.  
On the other hand, Prefix Tuning \citep{li-liang-2021-prefix} optimizes the inputs of every attention layer in the pre-trained LMs. 

Constructing soft visual prompts for a frozen LLM is an effective way to achieve vision-language alignment \citep{merullo2023linearly, pmlr-v202-koh23a}.
Flamingo \citep{alayrac2022flamingo} adds cross-attention layers to the LLM for incorporating visual features, pretrained on billions of image-text pairs. 
BLIP-2 \citep{li2023blip} adopts a Q-Former module to link a frozen image encoder to a frozen LLM, learning visual features relevant to text. LLaVA \citep{NEURIPS2023_6dcf277e},  trained on multimodal instruction-following, uses a linear layer to map image features from pre-trained CLIP to the word embedding space of Vicuna \citep{chiang2023vicuna}.
Inspired by the widespread application of visual prefix tuning in V\&L tasks, we explore its potential in visual storytelling while also considering the context when tuning the prefix.


\section{Method}
\label{sec:method}

\begin{figure*}[t!]
  \centering
\includegraphics[width=0.88\linewidth]{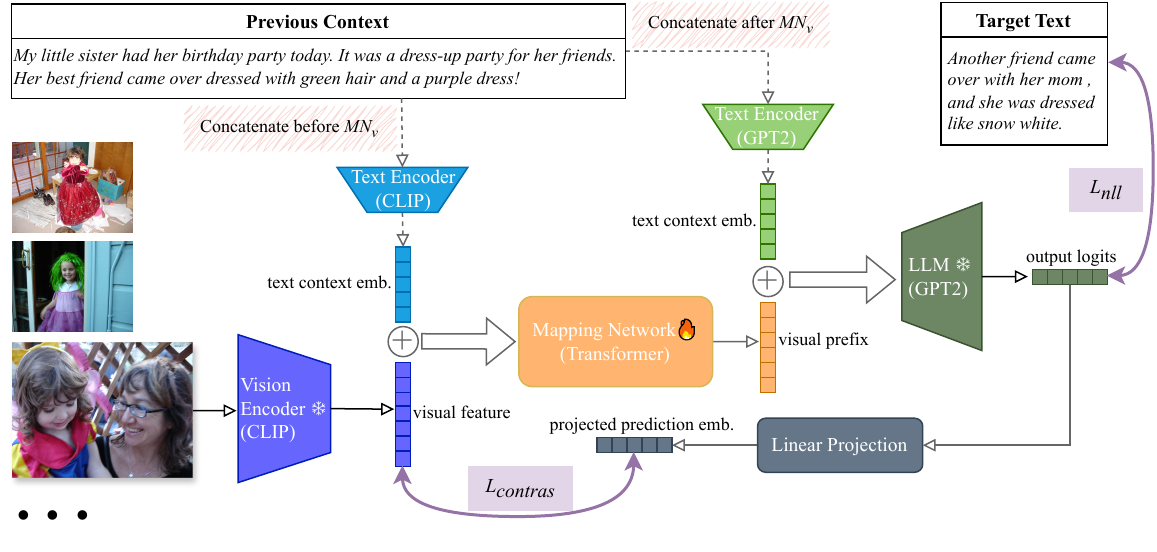}
  \caption{Illustration of the framework. A Transformer-based mapping network ($\mathcal{MN}_{\mathrm{v}}$) is trained to map visual features from a frozen encoder (CLIP) into a visual prefix for a frozen LLM (GPT2). We incorporate the previous sentences as the context via (1) concatenation after $\mathcal{MN}_{\mathrm{v}}$: previous context is encoded by the LLM (GPT2), combined with the visual prefix and then fed into the LLM decoder; or (2) concatenation before $\mathcal{MN}_{\mathrm{v}}$: previous context is encoded by the CLIP text encoder, combined with CLIP visual features and then fed into $\mathcal{MN}_{\mathrm{v}}$.
  In addition to the teacher-forcing objective $\mathcal{L}_{\mathrm{NLL}}$, we further compel the model to produce text that aligns semantically with the image through a contrastive training objective $\mathcal{L}_{\mathrm{contras}}$.
  }
  \label{fig:model}
\end{figure*}

In visual storytelling, the input  is a sequence of $N$ images $\mathcal{I}=\left\{I_1, \ldots, I_N\right\}$, where $N = 5$ in the VIST dataset \citep{huang_visual_2016}. Our model aims to generate a multi-sentence story $\mathcal{S}$ by predicting the probability $P(\mathcal{S}|\mathcal{I})$. In this 
section, we introduce a visual storytelling pipeline enhanced with prefix tuning (\cref{sec-clipcap}), then describe the context-aware components (\cref{sec-context}), curriculum training (\cref{sec:curriculum}) and finally the contrastive learning loss involved (\cref{sec:contras}).
\autoref{fig:model} illustrates an overview of our framework.

\subsection{Visual Storytelling with Prefix Tuning}
\label{sec-clipcap}

From the perspective of a single image, visual storytelling is very similar to image captioning, 
where an image-sentence pair $\{I_{i}, S_{i}\}$ is given.
Motivated by prefix tuning \citep{li-liang-2021-prefix}, ClipCap \citep{mokady2021clipcap} only updates the parameters of a lightweight Transformer-based mapping network during training to produce visual prefix vectors that can drive a pretrained frozen language model (LM) to generate text. 
ClipCap applies frozen CLIP \citep{radford2021learning} as vision encoder to extract visual features from the input image as $\boldsymbol{v}_i = f_{\mathrm{CLIP}}\left(I_i\right)$. 
The visual feature $\boldsymbol{v}_i $ is then processed by a trainable mapping network $\mathcal{MN}_{\mathrm{v}}$ to map the visual features to visual prefix vectors that are in the embedding space of the LM:
{\small
\begin{align*}
\mathbf{p}_{I_i} = [p_1, \dots, p_k] =\mathcal{MN}_{\mathrm{v}}(\boldsymbol{v}_i)=\mathcal{MN}_{\mathrm{v}}{(f_{\mathrm{CLIP}}\left(I_i\right))} 
\end{align*}
}
where $k$ denotes the prefix size and $\mathcal{MN}_{\mathrm{v}}$ is a Transformer with 8 multi-head self-attention layers with 8 heads each.
We then concatenate the visual prefix vectors $\mathbf{p}_{I_i}$ to the caption tokens $S_i = [s_1, s_2, ..., s_\ell]$, as
\begin{align*} 
\small
\mathbf{z}_{I_i} = [p_1, \dots, p_k; s_1, \dots, s_\ell ]
\end{align*}
where  ` $;$' denotes the concatenation.
During training, $\mathbf{z}_i$ is fed into the LM with a teacher-forcing objective in an auto-regressive manner. In other words, the mapping network $\mathcal{MN}_{\mathrm{v}}$ is trained using Negative Log-Likelihood (NLL) loss:
\begin{align*}
\small
\mathcal{L}_{\mathrm{NLL}}=- \sum_{j=1}^{\ell} \log p_\theta\left(s_j \mid p_1, \ldots, p_k; s_1, \ldots, s_{j-1}\right)
\end{align*}
where $\theta$ are the trainable parameters of the model.

\subsection{Context-aware Mapping Network}
\label{sec-context}
VIST story generation needs to establish 
informative connections between images in a sequence to bridge the potential visual/semantic gaps between them.
We incorporate contextual knowledge into our model in the form of past story sentences.
In addition to the image, we use the previous $L$ sentences  $[\mathcal{S}_{i-L}, \dots, \mathcal{S}_{i-1}]$ to generate the sentence for the current image $I_i$. 
For the first image $I_0$ in a sequence, we use the title and description of the belonging album\footnote{
\citet{huang_visual_2016} collected 10,117 Flickr albums that each contains 10 - 50 images. They asked human annotators to select 5 images of each album to form an image sequence, and write a story correspondingly. 
Album titles, descriptions and other metadata were provided in the original Flickr albums by the album owners.} as the textual context. We propose two methods to include the previous sentences\footnote{During training, we use the previous  ground-truth sentences as the context, while during inference the past predicted sentences are used instead. } as additional contextual information: (1) Concatenate 
$[\mathcal{S}_{i-L}, \dots, \mathcal{S}_{i-1}]$
with visual prefix vectors $\mathbf{p}_{I_i}$; (2) Concatenate 
$[\mathcal{S}_{i-L}, \dots, \mathcal{S}_{i-1}]$
with visual features $\boldsymbol{v}_i$ and use them together as the input of mapping network.

\paragraph{Concatenate after $\mathcal{MN}_{\mathrm{v}}$.}
Following \citet{han2023autoad}, we embed the sentences $[\mathcal{S}_{i-L}, \dots, \mathcal{S}_{i-1}]$ with the language generation model $f_{\mathrm{LM}}$ as 
{\small 
\begin{align*}
\mathbf{C}text_{i}=\left[\mathrm{BOS}_{\mathrm{text}} ; 
    f_{\mathrm{LM}} ([\mathcal{S}_{i-L}, \dots, \mathcal{S}_{i-1}]); \mathrm{EOS}_{\mathrm{text}}\right]
\end{align*}
}%
where $\mathrm{BOS}_{\mathrm{text}}$ and $\mathrm{EOS}_{\mathrm{text}}$ are learnable beginning and end of sequence tokens. The contextual vector $\mathbf{C}text_{i}$ is concatenated with the prefix vector $\mathbf{p}_{I_i}$ and then fed to the language generation model as a prompt vector (see Figure \ref{fig:model}). $\mathcal{MN}_{\mathrm{v}}$ is trained with NLL loss as: 
{\small
\begin{align*}
\mathcal{L}_{\mathrm{NLL}}=- \sum_{j=1}^{\ell} \log p_\theta\left(s_j \mid \mathbf{p}_{I_i} ; \mathbf{C}text_{i} ; s_1, \ldots, s_{j-1}\right)
\end{align*}
}%

\paragraph{Concatenate before $\mathcal{MN}_{\mathrm{v}}$.}
Since CLIP \citep{radford2021learning} is multimodal, we can use a common embedding space to encode both the image $I_i$ as $f_{\mathrm{CLIP}}\left(I_i\right)$, and previous sentences $[\mathcal{S}_{i-L}, \dots, \mathcal{S}_{i-1}]$ as $f_{\mathrm{CLIP}}([\mathcal{S}_{i-L}, \dots, \mathcal{S}_{i-1}])$. The two CLIP embeddings are then concatenated and fed into the mapping network to produce visual prefix vectors
\begin{align*}
\small 
\mathbf{p}'_{I_i} = \mathcal{MN}_{\mathrm{v}} ([f_{\mathrm{CLIP}}\left(I_i\right);f_{\mathrm{CLIP}}([\mathcal{S}_{i-L}, \dots, \mathcal{S}_{i-1}])]).
\end{align*}
The $\mathcal{MN}_{\mathrm{v}}$ is trained with NLL loss as:
\begin{align*}
\small
\mathcal{L}_{\mathrm{NLL}}=- \sum_{j=1}^{\ell} \log p_\theta\left(s_j \mid \mathbf{p}'_{I_i} ; s_1, \ldots, s_{j-1}\right)
\end{align*}

\subsection{Curriculum Learning}
\label{sec:curriculum}
In VIST, reference texts are often too generic and lack concretness to the image content. An example is "There was a lot to see and do" for an image depicting a funfair.
The frequency of this phenomenon may compromise the model's ability to ground its linguistic choices in visual data. 
To address this, we use curriculum learning, which involves training a model with data sorted by difficulty to improve generalization and speed up convergence \citep{bengio2009curriculum}.

We start by training the model on basic image captioning data to enhance grounding abilities before progressing to storytelling from image sequences. 
The training proceeds as follows: 
\textbf{(1)} Train the mapping network $\mathcal{MN}_{\mathrm{v}}$ with image-caption pairs (Description in Isolation, DII) from VIST (see Section \ref{sec:dataset}).
\textbf{(2)} Switch to visual storytelling data (Stories in Sequence, SIS) once validation loss stops decreasing.
\textbf{(3)} Return to step \textbf{(1)} when validation loss stops decreasing.
\textbf{(4)} Stop training when no further improvement in validation loss is observed.

\subsection{Visually-supervised Contrastive Training}
\label{sec:contras}

To encourage our model to generate text that is grounded in the image,
we leverage
a contrastive training objective $\mathcal{L}_{\mathrm{contras}}$ in addition to the teacher forcing objective $\mathcal{L}_{\mathrm{NLL}}$. 
To maximize the relatedness
between a positive pair consisting of a target text sequence and a source image, while minimizing the similarity between the negative pairs, we apply InfoNCE (Noise-Contrastive Estimation) loss \citep{oord2018representation} as:
\begin{align*}
    \small
    \mathcal{L}_{\text {contras}}=- \log \frac{\exp \left(\operatorname{sim}\left(\boldsymbol{v}_i, \hat{S_i}\right) / \tau\right)}{\sum_{j \neq i}^{|B|} \exp \left(\operatorname{sim}\left(\boldsymbol{v}_i, \hat{S_j}\right) / \tau\right)}
\end{align*}
where $\hat{S_i}$ is the projected representation of the text decoder's final layer output via a linear projection layer, $\operatorname{sim(,)}$ denotes the cosine similarity of the two vectors, $|B|$ is the batch size, and $\tau$ denotes the temperature.

During training, we first train the mapping network with the NLL loss $\mathcal{L}_{\text {NLL}}$ (training DII and SIS data in curriculum training scheme) for the first $N_{nll}$ epochs and then add the contrastive loss $\mathcal{L}_{\text {contras}}$ (using only SIS data).
The reason for not using $\mathcal{L}_{\text {contras}}$ from the beginning is that initially the model can only generate random tokens, which cannot be projected to semantically meaningful embeddings for contrasting with the image representation.
Overall, our model is trained by minimizing the combined loss $\mathcal{L}$ \citep{zhu-etal-2023-visualize} as:
\begin{align*}
    \small
    \mathcal{L}= \begin{cases}\mathcal{L}_{\text {NLL}}, epoch < N_{nll}\\\mathcal{L}_{\text {NLL}}+\lambda \mathcal{L}_{\text {contras}}, epoch \geq N_{nll} \end{cases}
\end{align*}
where $\lambda$ is the coefficient of the contrastive loss.

\section{Experiments\protect\footnote{Experimental details of training, inference and automatic evaluation are listed in the \autoref{sec:append-exp}.}}

\begin{table}[t!]
\centering
\small
\begin{tabular}{@{}llll@{}}
\toprule
      &                                       & Original & Ours   \\ \midrule
Train & \multicolumn{1}{l|}{No. DII captions} &  120,465 & 120,099     \\
      & \multicolumn{1}{l|}{No. SIS stories\tablefootnote{Each story usually consists of 5 sequences of text corresponding to 5 images.}}  & 40,098   & 40,071 \\ \midrule
Val   & \multicolumn{1}{l|}{No. DII captions} & 14,970         & 14,940       \\
      & \multicolumn{1}{l|}{No. SIS stories}      & 4,988    & 4,988  \\ \midrule
Test  & \multicolumn{1}{l|}{No. DII captions} & 15,165   & 15,165       \\
      & \multicolumn{1}{l|}{No. SIS stories}      & 5,050    & 5,030  \\ \bottomrule
\end{tabular}
\caption{Data split in original VIST dataset annotations and our experiments. Differences are due to the removal of unavailable images for some samples. DII: Descriptions of Images in Isolation. SIS: Stories of Images in Sequence.}
\label{tab:data-split}
\end{table}

\subsection{Dataset}\label{sec:dataset}
The visual storytelling \cite[VIST;][]{huang_visual_2016} dataset includes 210,819 unique photos and 50,200 stories collected from 10,117 Flickr albums. 
Our experiments follow the data splits  in the original VIST, removing the broken or unavailable image files (see \autoref{tab:data-split}). 



\subsection{Decoding Strategies}
We compare four popular decoding methods for text generation:
\textbf{Beam search} selects the text continuation with highest probability based on the model’s probability distribution; this may result in low variation
\citep{li-etal-2016-diversity} and degeneration \citep{fan-etal-2018-hierarchical, holtzman2019curious} in the generated text. 
\textbf{Top-$k$ sampling} redistributes the probability mass among only the top
$k$ most likely next tokens, 
avoiding sampling from the unreliable tail of the distribution \citep{fan-etal-2018-hierarchical}. 
\textbf{Nucleus sampling} \citep{holtzman2019curious}, also known as top-$p$ sampling, chooses from the smallest set of tokens
whose cumulative probability exceeds the probability $p$.
\textbf{Contrastive search} \citep[SimCTG,][]{NEURIPS2022_871cae8f} jointly considers the probability predicted by the language model and the similarity with respect to the previous context. 

\subsection{Baseline Models}
\label{sec:baselines}
For a fair and thorough comparison, we choose four SOTA baselines that don't require additional datasets and have reproducible code/weights. \textbf{GLACNet} \citep{taehyeong_kim_glac_2018} is a seq2seq model using global-local attention and context cascading on visual features. \textbf{AREL} \citep{wang_no_2018} is an adversarial framework learning an implicit reward function from human demonstrations and optimizing policy search with a CNN-based reward model. \textbf{ReCo-RL} \citep{hu_what_2020} is a reinforcement learning model with composite rewards for relevance, coherence, and expressiveness. \textbf{TAPM} \citep{yu_transitional_2021} uses an adaptation loss to align a vision encoder with a pretrained LM and a sequential coherence loss to improve temporal coherence by aligning predicted text representations with neighboring visual representations.

\subsection{Automatic Evaluation Metrics}
In line with prior work on the VIST benchmark, we validate our results over the test set using the standard metrics BLEU \citep{papineni-etal-2002-bleu}, ROUGE-L \citep{DBLP:conf/acl/LinO04}, METEOR \citep{banerjee-lavie-2005-meteor}, CIDEr \citep{Vedantam_2015_CVPR} and SPICE \citep{DBLP:conf/eccv/AndersonFJG16}. 
We evaluate the generated text in terms of text-text semantic similarity using BLEURT
\citep{sellam-etal-2020-bleurt}, image-text semantic similarity using CLIPScore
\citep{hessel-etal-2021-clipscore}, and language fluency using Perplexity.
Following \citet{NEURIPS2022_871cae8f}, we also assess text degeneration and word diversity using:
(1) rep-$n=1.0-\frac{\mid \text { unique } n \text {-grams } \mid}{\mid \text { total } n \text {-grams } \mid}$ measures story-level repetition by computing the portion of duplicate $n$-grams; (2) diversity$=\prod_{n=2}^4(1-$ rep-$n)$ measures the diversity
of $n$-grams.

\subsection{Human Evaluation}
We conduct a human evaluation on a sample of generated texts.
We randomly select 100 distinct image sequences and the corresponding generated stories from 8 models (i.e., our model\footnote{We choose GPT2-xl, concatenation before mapping network, with curriculum learning and contrastive training, 
based on automatic metrics%
.} with four decoding strategies, the ground truth texts (GT), GLACNet, AREL and TAPM). 

We invite 75 human annotators from Prolific to rate stories on a 5-point Likert scale for the criteria of \textbf{Visual Grounding}, \textbf{Coherence}, \textbf{Interestingness}, and \textbf{Informativeness}. As noted in Section~\ref{sec:intro}, we consider these among the most important criteria for visually grounded narrative generation.
Each participant answered 32 questions (each question containing ratings for one image sequence and one story across four criteria), resulting in a total of 9600 responses.
We evenly distributed 800 pairs of image sequences and stories among all participants, ensuring that each question received 
$\sim$3
responses. 
A full explanation of rating criteria, questionnaire instructions and sample questions are in the \autoref{sec:append-survey}.

\section{Results and Analysis}

\begin{table}[!h]
\centering
\small
\setlength{\tabcolsep}{3pt}
\begin{tabular}{lccccccc}
\toprule
\multicolumn{1}{l}{Setting}        & \textbf{B-4} & \textbf{M}   & \textbf{R-L}  &\textbf{C}    &\textbf{S}    & \textbf{BR}   & \textbf{PPL}$\downarrow$  \\ \midrule
\multicolumn{1}{l|}{GLACNet} & 13.5 & 31.6 & \textbf{30.0} & 7.6  & 8.3  & 30.7 & 12.0 \\
\multicolumn{1}{l|}{AREL}    & 13.5 & \textbf{31.7} & 29.6 & 8.6  & 8.9  & 30.4 & 13.1 \\
\multicolumn{1}{l|}{TAPM}    & 11.4 & 30.7 & 28.7 & 9.5  & 10.0 & 31.4 & 18.3 \\
\multicolumn{1}{l|}{ReCo-RL} & 13.1 & 31.5 & 27.9 & 11.5 & \textbf{11.2} & 27.7 & 28.4 \\ \midrule
\multicolumn{8}{l}{\textit{no context}}                                                \\
\multicolumn{1}{l|}{beam}    & 9.8  & 27.4 & 27.2 & 5.0  & 5.9  & 26.7 & 13.9 \\
\multicolumn{1}{l|}{top-$k$} & 4.0  & 24.1 & 22.5 & 2.1  & 6.6  & 24.9 & 39.7 \\
\multicolumn{1}{l|}{nucleus} & 3.5  & 23.6 & 21.4 & 1.7  & 5.7  & 24.1 & 42.5 \\
\multicolumn{1}{l|}{SimCTG}  & 7.3  & 28.5 & 25.5 & 5.7  & 6.9  & 25.8 & 16.6 \\ \midrule
\multicolumn{8}{l}{\textit{+context after} $\mathcal{MN}_{\mathrm{v}}$}                                        \\
\multicolumn{1}{l|}{beam}    & 13.6 & 31.4 & 29.0 & 11.4 & 9.7  & 31.5 & \textbf{10.5} \\
\multicolumn{1}{l|}{top-$k$} & 4.0  & 25.1 & 22.4 & 5.8  & 8.9  & 29.1 & 32.9 \\
\multicolumn{1}{l|}{nucleus} & 3.5  & 24.2 & 22.0 & 5.6  & 7.9  & 28.2 & 41.6 \\
\multicolumn{1}{l|}{SimCTG}  & 7.9  & 28.8 & 26.0 & 7.5  & 9.7  & 30.6 & 13.3 \\ \midrule
\multicolumn{8}{l}{\textit{+context before} $\mathcal{MN}_{\mathrm{v}}$}                                         \\
\multicolumn{1}{l|}{beam}    & \textbf{14.0} & 31.2 & 29.3 & \textbf{12.0} & 9.9  & \textbf{32.4} & 11.1 \\
\multicolumn{1}{l|}{top-$k$} & 4.9  & 25.1 & 23.5 & 5.8  & 7.9  & 28.3 & 33.2 \\
\multicolumn{1}{l|}{nucleus} & 4.2  & 24.0 & 22.78 & 5.5  & 7.4  & 27.2 & 42.2 \\
\multicolumn{1}{l|}{SimCTG}  & 7.7  & 29.0 & 26.1 & 7.6  & 8.4  & 30.9 & 12.7 \\ \bottomrule
\end{tabular}
  \caption{Automatic evaluation results on VIST test set. All listed  models are trained with curriculum learning and contrastive loss using GPT2-xl as language generator. B-4: BLEU-4; M: METEOR; R-L: ROUGE-L; C: CIDEr; S: SPICE; BR: BLEURT; PPL: Perplexity.}
\label{tab:results-main}
\end{table}

\subsection{Automatic Evaluation}
\label{sec:auto-eval}

Table \ref{tab:results-main} outlines the results of automatic metrics among the baselines\footnote{Following the original papers, all the baselines use beam search as decoding strategy.} and our models with curriculum learning, contrastive training and 
GPT2-xl as the decoder (we consider the impact of different decoder model sizes further below).
These results suggest that our model is comparable to or 
better than the strong baselines on most automatic metrics.

In our experiments, we found that using or not using curriculum learning has no significant impact on automatic metrics (see the full report in the \autoref{sec:append-result}). 
In what follows, we will specifically analyze the impact of the textual context, contrastive training, language model size, and decoding strategies on our method, plus the evaluation of linguistic diversity.

\paragraph{Textual context.}

\begin{figure}[t!]
  \centering
  \begin{minipage}[b]{0.75\linewidth}
    \includegraphics[width=\linewidth]{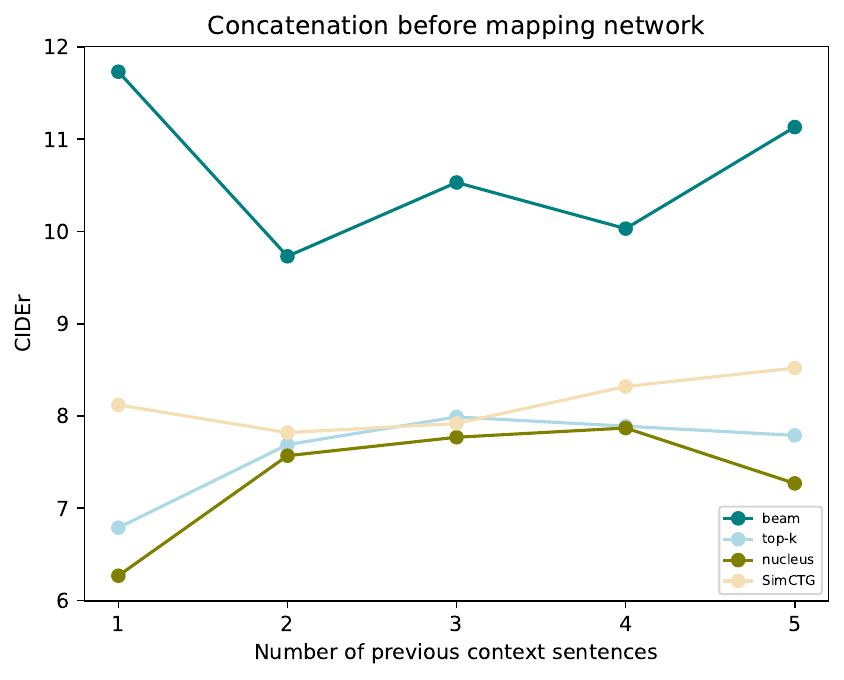}
  \end{minipage}
  \hfill 
  \begin{minipage}[b]{0.75\linewidth}
    \includegraphics[width=\linewidth]{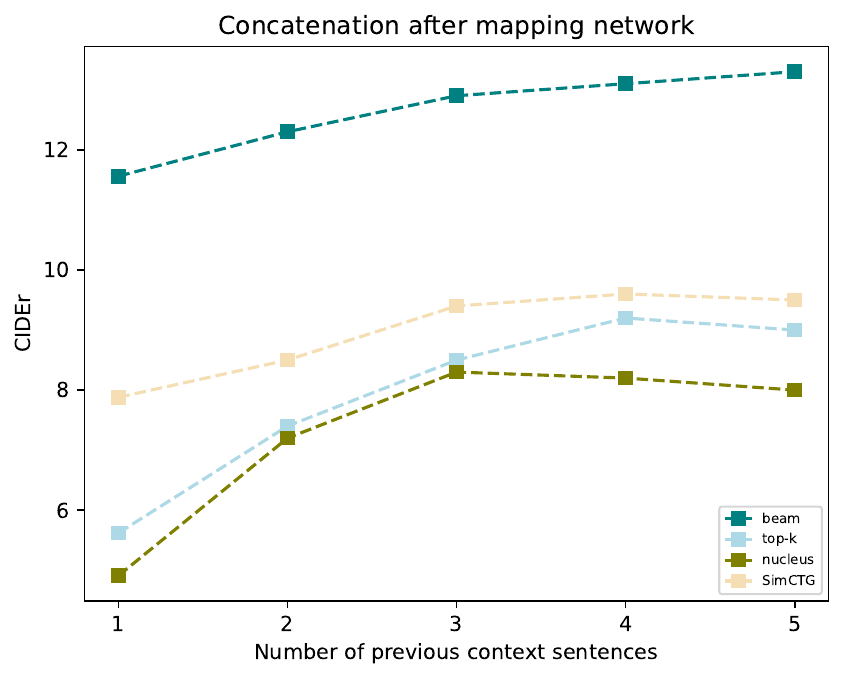} 
  \end{minipage}
  \caption{Impact of context length: CIDEr of various number of previous context sentences with concatenation before (top) and after (bottom) $\mathcal{MN}_{\mathrm{v}}$.}
  \label{fig:concat_nums}
\end{figure}

Table \ref{tab:results-main} demonstrates that the combination of textual context (num of previous sentences = 1) brings a consistent improvement, both when concatenation is before and after $\mathcal{MN}_{\mathrm{v}}$.
The third and the fourth blocks of \autoref{tab:results-main} show that the choice of concatenation strategy does not have much impact on the performance.

Figure \ref{fig:concat_nums} shows the impact of concatenating different numbers of previous sentences as context, in both settings.
For concatenation before $\mathcal{MN}_{\mathrm{v}}$ (top in Figure \ref{fig:concat_nums}), we observe that performance tends to decline as context gets longer when decoding with beam search and contrastive search.
Whereas, the performance slightly improves for top-k and nucleus sampling when the number of context sentences is less than 3 and 4, respectively.
This may be due to the restriction of the maximum length of the input to CLIP to 77 tokens
\footnote{When the previous context length exceeds 77 tokens, we discard the excess.}.
For the context concatenation after $\mathcal{MN}_{\mathrm{v}}$ (bottom in Figure \ref{fig:concat_nums}), extending the context length marginally enhances performance, yet it also  incurs additional computational costs because of the quadratic complexity of the attention mechanism in GPT2.

\paragraph{Contrastive training.}

\begin{figure}[t!]
  \centering
  \begin{minipage}[b]{0.75\linewidth}
    \includegraphics[width=\linewidth]{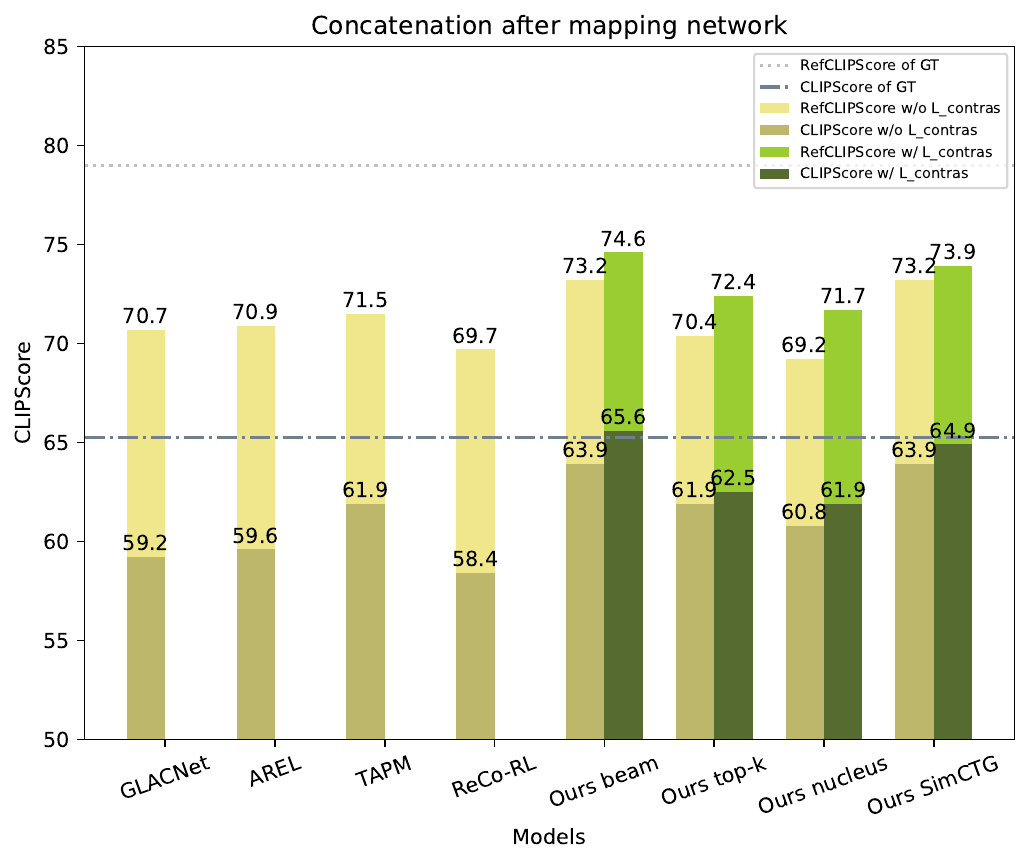}
  \end{minipage}
  \hfill 
  \begin{minipage}[b]{0.75\linewidth}
    \includegraphics[width=\linewidth]{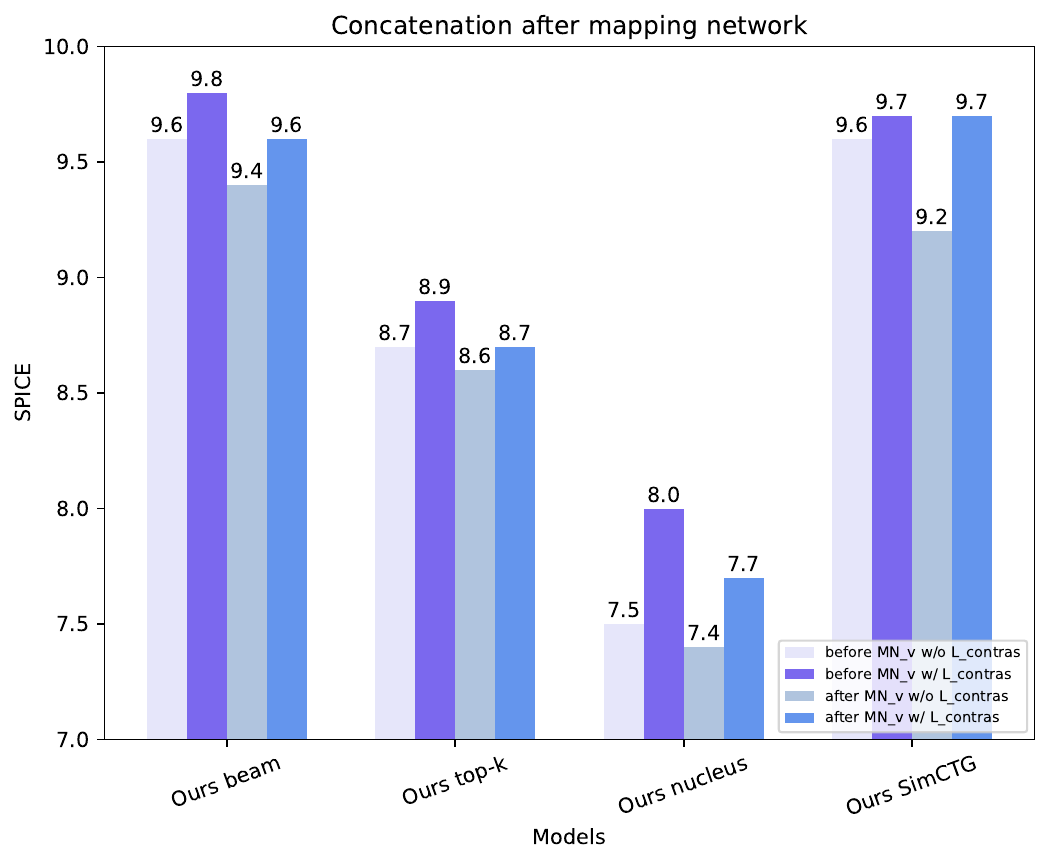} 
  \end{minipage}
  \caption{Impact of contrastive training object: CLIPScore (top) and SPICE (bottom) of training our models without or with $\mathcal{L}_{\text {contras}}$ .}
  \label{fig:contras_training}
\end{figure}

We explore the impact of the contrastive training objective  with CLIPScore and RefCLIPScore \citep{hessel-etal-2021-clipscore} shown on the top of Figure \ref{fig:contras_training}. Contrastive training brings about a clear gain for both CLIPScore and RefCLIPScore, as the contrastive loss serves to minimize the difference between the generated text and the image content in the semantic space of CLIP. 
In addition to the improvement of text-image similarity, incorporating 
$\mathcal{L}_{\text{contras}}$ also produces higher SPICE scores, as shown on the bottom of Figure \ref{fig:contras_training}. 
This implies that stories generated with contrastive training are more semantically accurate and detailed, effectively describing important elements and their interrelations in the images.

\begin{figure}[]
  \centering
  \begin{minipage}[b]{0.75\linewidth}
    \includegraphics[width=\linewidth]{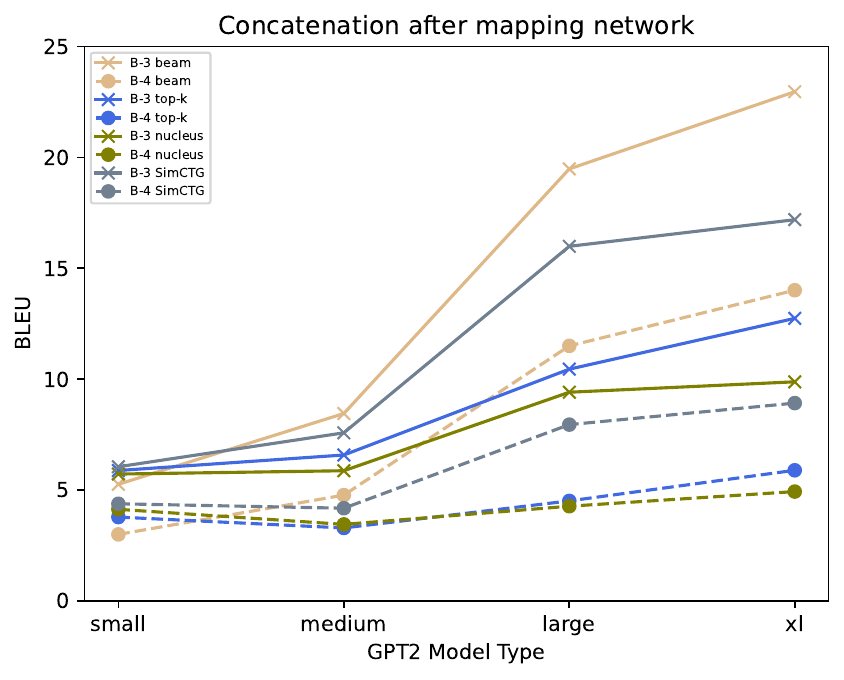}
  \end{minipage}
  \hfill 
  \begin{minipage}[b]{0.75\linewidth}
    \includegraphics[width=\linewidth]{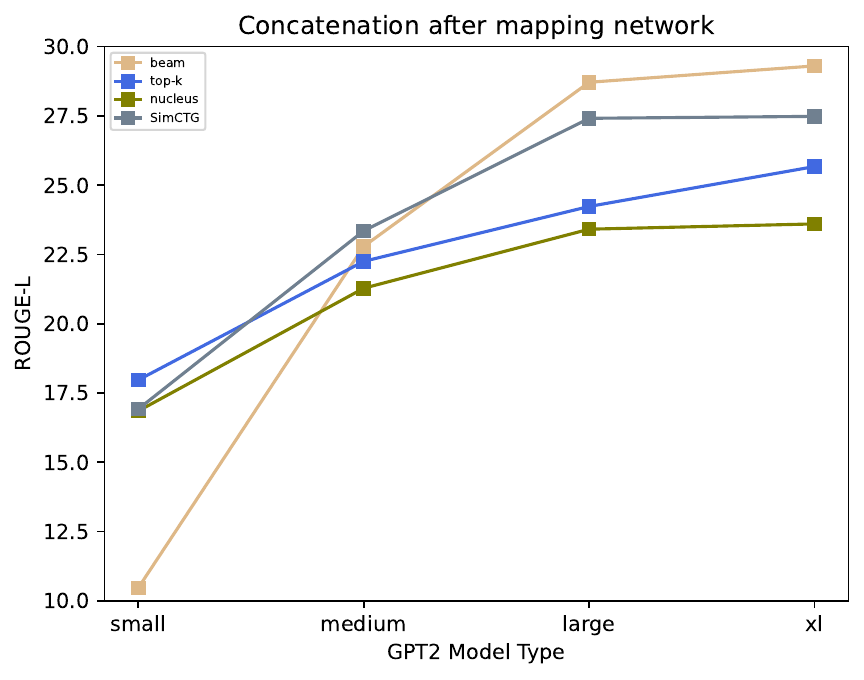} 
  \end{minipage}
  \caption{Impact of language model size: BLEU-3, 4 (top) and ROUGE-L (bottom) of our models using GPT2-small, medium, large and xl as text generator with textual context concatenation after $\mathcal{MN}_{\mathrm{v}}$.}
  \label{fig:lm-size}
\end{figure}

\paragraph{Language model size.}
Figure \ref{fig:lm-size} illustrates the performance of various decoding methods applied to different sizes of the GPT2 model.
As the model size increases, all decoding methods tend to yield higher BLEU and ROUGE-L scores, especially when comparing GPT2-small to GPT2-large, with limited additional benefits accrued from the larger GPT2-xl.
Full results of different language models are in \autoref{sec:append-result}.

\paragraph{Decoding strategies.}
Under identical training, different decoding methods exhibit varying performance across various automatic metrics (as shown in Table \ref{tab:results-main}, Figures \ref{fig:concat_nums}, \ref{fig:contras_training}, \ref{fig:lm-size}).
Beam search performs the best among all automatic metrics followed by SimCTG, while top-$k$ and nucleus sampling 
score worse.
Though beam search suffers from high repetition and yields very generic text,
it seems to align better with the ground truth based on standard automatic metrics in image captioning. 
On the other hand, decoding methods that aim at alleviating text degeneration, like top-$k$ and nucleus sampling, tend to generate stories that differ from the ground truth, perhaps due to 
hallucination.
SimCTG seems to strike a better balance between grounding and degeneration for VIST. 
These somewhat counter-intuitive results provide the strongest motivation for our human evaluation, which does not rely on a metric-based comparison of generated text to ground- truth narratives.

\begin{table}[t!]
\small
\centering
\setlength{\tabcolsep}{1pt}
\begin{tabular}{@{}cccccc@{}}
\toprule
                             & \textbf{rep-1}$\downarrow$  & \textbf{rep-2}$\downarrow$ & \textbf{rep-3}$\downarrow$ & \textbf{rep-4}$\downarrow$ & \textbf{diversity}$\uparrow$ \\ \midrule
\multicolumn{1}{c|}{GT}      & 26.94  & 4.22 & 1.03 & 0.39 & 94.43     \\ \midrule
\multicolumn{1}{c|}{GLACNet}      & 48.43  & 27.77 & 20.86 & 15.97 & 48.03     \\
\multicolumn{1}{c|}{AREL}    & 45.20   & 22.04 & 15.16 & 10.98 & 58.88     \\
\multicolumn{1}{c|}{TAPM}    & 36.16  & 10.02 & 5.16 & 2.89 & 82.87     \\
\multicolumn{1}{c|}{ReCo-RL}    & 33.58 & 3.14 & \textbf{0.11} & \textbf{0.02} & 97.27     \\ \midrule
\multicolumn{6}{c}{\tiny Concatenate \textbf{before} $\mathcal{MN}_{\mathrm{v}}$, \textbf{without} contrastive training, GPT2-xl}              \\ \midrule
\multicolumn{1}{c|}{beam}    & 55.33  & 37.22 & 29.49 & 23.91 & 33.68     \\
\multicolumn{1}{c|}{top-$k$}    & 26.80   & 2.80  & 0.39 & 0.08 & 96.74     \\
\multicolumn{1}{c|}{nucleus} & 24.72  & 2.07 & 0.23 & 0.05 & 97.64     \\
\multicolumn{1}{c|}{SimCTG}   & 35.02  & 8.53 & 2.53 & 0.89 & 88.36     \\ \midrule
\multicolumn{6}{c}{\tiny Concatenate \textbf{before} $\mathcal{MN}_{\mathrm{v}}$, \textbf{with} contrastive training, GPT2-xl}                  \\ \midrule
\multicolumn{1}{c|}{beam}    & 48.31  & 26.18 & 18.32 & 13.38 & 52.23     \\
\multicolumn{1}{c|}{top-$k$}    & 26.55  & 2.67 & 0.36 & 0.08 & 96.91     \\
\multicolumn{1}{c|}{nucleus} & \textbf{24.40}   & \textbf{2.04} & 0.27 & 0.06 & \textbf{97.69}     \\
\multicolumn{1}{c|}{SimCTG}   & 33.16  & 7.18 & 1.87 & 0.61 & 90.53     \\ \bottomrule
\end{tabular}
\caption{Text degeneration analysis with rep-1,2,3,4 and diversity score.}
\label{tab:results-div}
\end{table}

\paragraph{Linguistic diversity assessment.}
The diversity metrics in Table \ref{tab:results-div} show that beam search suffers from severe text degeneration and `stammering', that is, generating repeated sequences.
In contrast, our models with nucleus sampling provide the most diverse expressions.
As shown in the second and third blocks in Table \ref{tab:results-div}, training our model with contrastive loss can also alleviate the degeneration problem with beam search decoding.
This further supports the effectiveness of contrastive training in reducing repetitive text.

\begin{table}[t!]
\centering
\small
\setlength{\tabcolsep}{1pt}
\begin{tabular}{@{}c|cccc@{}}
\toprule
& \textbf{\tiny Visual Grounding} & \textbf{\tiny Coherence}     & \textbf{\tiny Interestingness} & \textbf{\tiny Informativeness} \\ \midrule
\tiny GT                                                               & 4.10             & 3.71          & 3.10            & 3.61            \\ \midrule
\tiny GLACNet                                                          & 2.75             & 2.19          & 1.78            & 2.06            \\
\tiny AREL                                                             & 2.85             & 2.26          & 1.83            & 2.20            \\
\tiny TAPM                                                             & 3.16             & 2.82          & 2.34            & 2.61            \\ \midrule
\tiny Ours beam                                                             & 2.95             & 2.11          & 1.80            & 2.17            \\
\tiny Ours top-$k$                                                            & 3.01             & 2.57          & \textbf{2.40}   & 2.67            \\
\tiny Ours nucleus                                                          & 2.72             & 2.42          & 2.27            & 2.41            \\
\tiny Ours SimCTG                                                           & \textbf{3.20}    & \textbf{2.85} & 2.27            & \textbf{2.68}   \\ \midrule

\tiny $F$(6,293)                                                          &  6.38   &18.46 &    19.05         &  15.30 \\
\tiny $p$-value  & 1.16e-6         & 6.17e-21      & 1.22e-21        & 3.45e-17                                                    \\ \bottomrule
\end{tabular}
\caption{Human evaluation results: mean rating scores for ground truth (GT), baselines and our models, plus $F$-statistic and $p$-value of a one-way ANOVA comparing models on each evaluation dimension.}
\label{results-human}
\end{table}

\subsection{Human Evaluation}
\label{sec:human-eval}
Table \ref{results-human} displays the means of human rating scores for ground truth (GT), GLACNet, AREL, TAPM and our model with four decoding methods. 

Our model with SimCTG decoding outperforms other approaches in terms of Visual Grounding, Coherence and Informativeness. 
Our model with top-$k$ performs the best in Interestingness.
Thus, stories generated by our model compare favorably to baselines in human evaluation.
Crucially, we observe a strong discrepancy between the human evaluation results and automatic metrics. 
In particular, our model with beam search decoding is ranked low on human judgments, whereas it tends to be ranked highly on automatic metrics, especially those relying on a token-matching comparison to the reference texts.
A one-way ANOVA (see the last row of Table \ref{results-human}) shows that overall, differences between models on each of the human evaluation criteria are significant. We follow this up with pairwise comparisons using Tukey's HSD to identify the significant pairwise differences; see the \autoref{sec:append-tukey} 
for complete results. 
We find that
our model significantly outperforms GLACNet and AREL on human judgments, but is not statistically distinguishable from the other models, 
though our model leads in terms of mean values.



\begin{figure}[t!]
  \centering
\includegraphics[width=\linewidth]{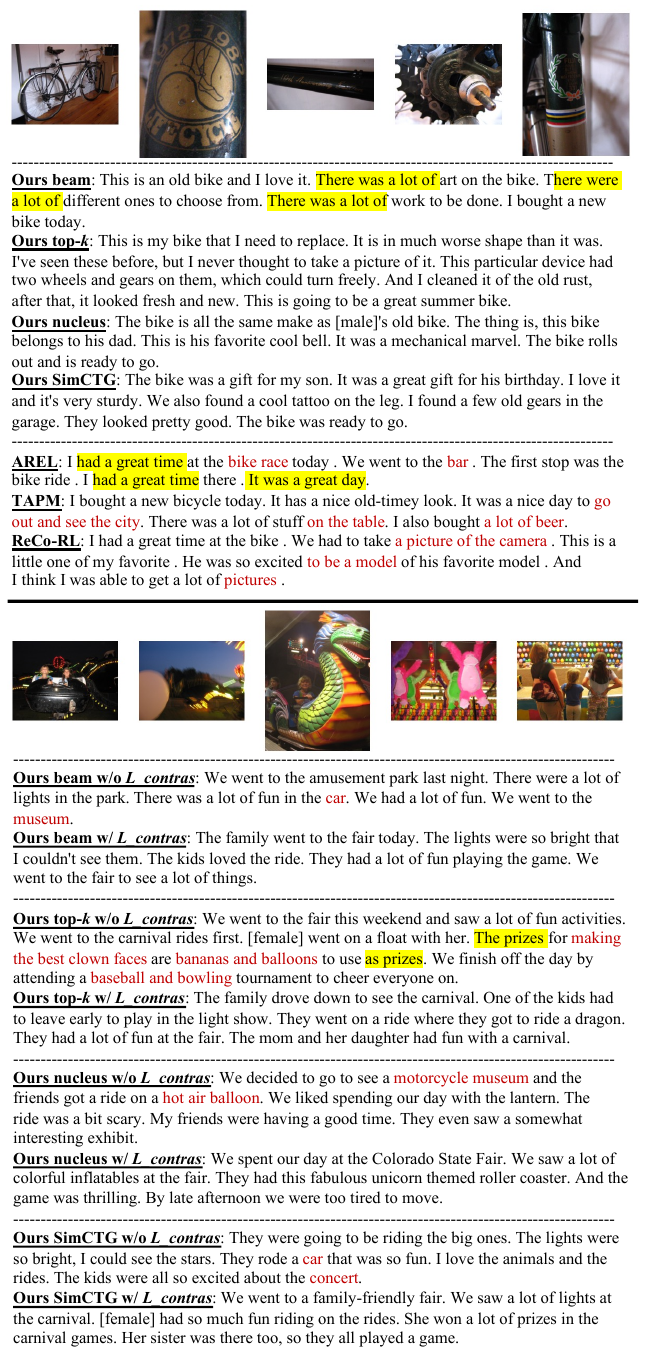}
  \caption{Qualitative examples of our model and baselines. Words highlighted in yellow are repetitive expressions, and words in red represent content that is not relevant to the image sequence.}
  \label{fig:case-all}
\end{figure}


\subsection{Qualitative Case Analysis}
\label{sec:case_analysis}

The first image sequence in \autoref{fig:case-all} shows stories generated by our models and the baselines 
in which our model's productions better ground to the input images involving an old bicycle and its various parts.
For example, TAPM includes unrelated expressions such as "\textit{on the table}" and "\textit{a lot of beer}", and ReCo-RL erroneously mentions "\textit{a picture of the camera}" and "\textit{to be a model}".
In contrast, our model consistently generates text closely relevant to the theme of "\textit{bike}" and provides more specific details, avoiding generic expressions like "\textit{had a great time}".
Our model demonstrates a promising ability to write coherent narratives with clear storylines, even for the challenging image sequence in the upper part of Figure \ref{fig:case-all} (where all images are of objects).
For instance, the story from our model with top-$k$ decoding features a fairly clear narrative arc, wherein the narrator discovers a bike in poor condition that is restored 
after repair and cleaning.
This further confirms our model's ability to generate more 
relevant and engaging stories.

The second image sequence of \autoref{fig:case-all} compares the stories generated by our models without and with contrastive training.
The contrastive training forces the model to generate more visually grounded stories with fewer irrelevant elements, that is, hallucinations. However, defining hallucinations in open-ended generation tasks like VIST remains challenging. While hallucinations can disrupt the story-image correspondence, they can also create intriguing narratives. The storytelling based on images is expected to incorporate elements which are not strictly descriptive of visual contents. For example, the last sentence in the story by our model with top-$k$ decoding and contrastive training, "\textit{We finish off the day by attending a baseball and bowling tournament to cheer everyone on}" is not directly reflected in the images but adds relevant context and imaginative extension. Balancing hallucination and creativity is left for future work.



\section{Conclusion}

We present a simple yet effective framework for visual storytelling that utilizes pretrained multimodal models with a lightweight vision-language mapping network to construct prefixes for LLMs.  
Our model enhances the coherence of multi-sentence stories by integrating contextual information.
In addition to teacher-forcing loss, we use a curriculum training scheme and image-text contrastive loss to enhance the concretness and visual grounding of generated stories.
Extensive evaluation on the VIST benchmark  using both automatic metrics and human assessment shows that our model obtains strong results compared to SOTA methods.
We empirically confirm that our model demonstrates the ability to generate coherent stories that are closely tied to visual content, and possess more creative and engaging details with minimal degeneration.
Our study contributes to improved evaluation practices in text generation, recommending a specific human evaluation setup for visual storytelling that assesses four key output qualities. 
Such evaluation enables informative model comparisons and better insight into the relative strengths of different systems. 
Results show that automatic metrics, particularly token overlap measures like BLEU, often poorly correspond to human judgments and should not be fully trusted for open-ended tasks like visual storytelling. This echoes similar observations made in other NLG domains~\cite{Belz2006,Reiter2009a,reiter_structured_2018,moramarco_human_2022}.

\paragraph*{Limitations.} 
Despite having employed diverse automatic metrics and comprehensive human evaluations to assess our models' generated stories, we recognize substantial opportunity for enhancing the evaluation methodology of visual storytelling.
As discussed above, correlating with ground-truth text or grounding to the visual content represents just a one-sided view, which 
downplays the role of diversity and creativity in storytelling. 
While our proposed human evaluation aims for thorough assessment, human annotation is costly and cannot be continuously applied during model development.
Future research could explore the balance in visual storytelling between factuality and groundedness on the one hand, and {\em justified} deviation from the images in the interest of creativity on the other. 

Additionally, our model exhibits certain biases, such as producing wedding-related stories from images of churches, even though there are no wedding-related elements in the images. This may stem from the biases in VIST dataset or the pretraining data of CLIP and GPT2.

Lastly, this study primarily investigates the utility and performance of two specific pre-trained models, CLIP and GPT-2. While these models have demonstrated broad applicability and strong performance across various tasks, they represent only a subset of the rapidly evolving landscape of pre-trained vision an language models. Future work could benefit from incorporating a wider array of models, such as BLIP-2 \citep{li2023blip}, LLaVA \citep{NEURIPS2023_6dcf277e}, Llama 3 \citep{meta2024llama} and Mistral \citep{mistral2024capabilities}, to provide a more comprehensive understanding of the strengths and limitations inherent to different foundation models.


\paragraph*{Ethics Statement.}
In this research, we employ pretrained multimodal models LLMs to transform images into narratives. There's a possibility that any biases inherent in the pre-training data may unintentionally be reflected in the text generated, potentially leading to uncontrolled biases. 
While our examination did not observe such problems, we recognize it as a potential concern that might affect the integrity of the generated content. 
Regarding the VIST dataset and the models used in this study, we are not aware of any major ethical concerns they may pose on their own. However, we acknowledge the potential for biases present in the original VIST data to influence both our models and their evaluations. Our research has received approval from the Ethics Board of our institution, ensuring compliance with ethical standards in human evaluation processes. All the human evaluation data collected has been de-identified to protect the privacy and security of all participants involved. 

\section*{Acknowledgements}
We thank the three anonymous reviewers for helpful comments, and the participants in our human evaluation for their role in evaluating our systems.




\bibliography{inlg2024}
\clearpage
\appendix

\section{Experimental Details of Training, Inference and Automatic Evaluation}
\label{sec:append-exp}
We use CLIP RN50x4 as the image encoder backbone to extract visual features offline\footnote{We tried both CLIP RN50x4 and CLIP ViT/B-32 in the preliminary experiments, and RN50x4 performs a little bit better than ViT/B-32. } and GPT2-small, medium, large and xl as the language decoder. The mapping network is a Transformer-based model with 8 multi-head self-attention layers with 8 heads each. We set the CLIP embedding length as 20 and visual prefix length as 20. We stop the text generation when an end of sequence token is predicted, otherwise we limit the maximum length to 30 tokens. 
For each experiment, we use a single NVIDIA A100 for training and inference. Other empirically tuned hyperparameters are listed in the Table \ref{tab:hyper-para}.

\begin{table}[h!]
\centering
\begin{tabular}{@{}l|l@{}}
\toprule
Hyperparameters & Value \\ \midrule
Batch size      & 50    \\
Training epochs & 10    \\
$N_{nll}$     & 6     \\
$\lambda$       & 0.3   \\
Optimizer       & Adam  \\
Learning rate   & 2e-5  \\
Weight decay    & 1e-4  \\
Warmup steps    & 1300  \\  \midrule
Max length & 30 \\
Num of beams & 5 \\
$k$ in top-$k$ & 50 \\
$p$ in nucleus sampling & 0.9 \\
Top-$k$ in SimCTG & 5 \\
Degeneration penalty in SimCTG & 0.8 \\
Temperature & 1.0 \\
\bottomrule
\end{tabular}
\caption{Hyperparameter settings.}
\label{tab:hyper-para}
\end{table}

As for the automatic evaluation, we use pycocoevalcap\footnote{https://github.com/tylin/coco-caption} library to compute BLEU, ROUGE-L, CIDEr and SPICE, and use the official VIST challenge evaluation code\footnote{https://github.com/windx0303/VIST-Challenge-NAACL-2018} to compute METEOR. 
We report BLEURT\footnote{https://github.com/google-research/bleurt} score with BLEURT-20 as the checkpoint, CLIPScore and RefCLIPScore\footnote{https://github.com/jmhessel/clipscore} with ViT-B/32 as the base model, and the mean perplexity\footnote{https://huggingface.co/spaces/evaluate-metric/perplexity} score calculated by GPT2.

\section{Human Evaluation Survey}
\label{sec:append-survey}
For the human evaluation survey, participants were asked to rate each pair, consisting of a story and an image sequence, on the following criteria: 
(1) \textbf{Visual Grounding} assesses how accurately and reasonably the story corresponds to the content in the image sequence; 
(2) \textbf{Coherence} evaluates how logical and consistent the story is; (3) \textbf{Interestingness} measures how the story captures the reader’s interest through unique ideas or expressions; 
(4) \textbf{Informativeness} evaluates how specific and detailed the story is in narrating the scene, objects, and events depicted in the images, rather than relying on highly generic
descriptions.

\begin{figure*}[h!]
  \centering
\includegraphics[width=\linewidth]{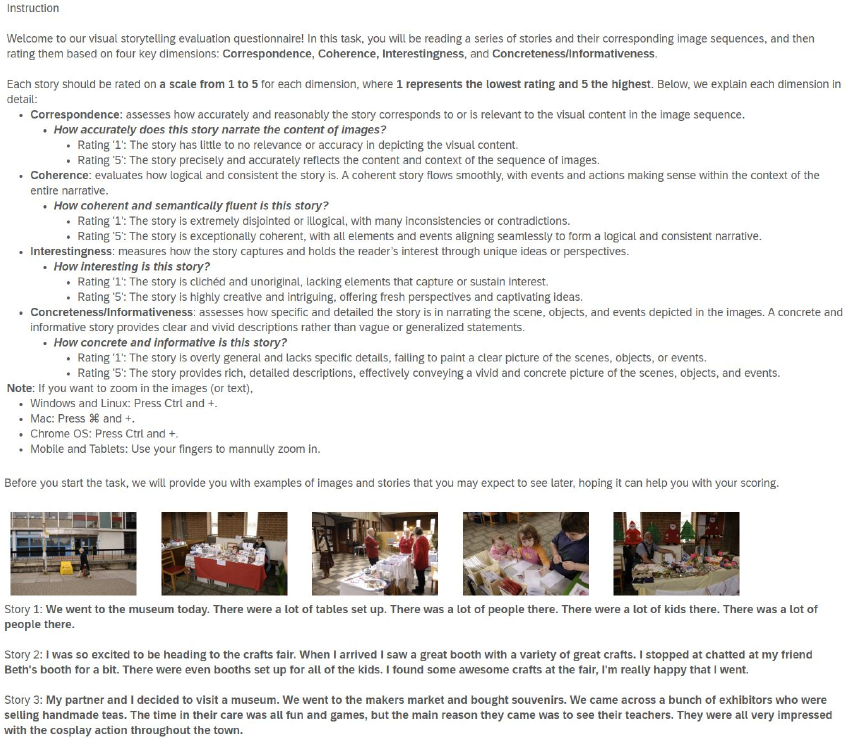}
  \caption{ Instructions, sample image sequence and corresponding stories we displayed at the beginning of the human evaluation questionnaire.
  }
  \label{fig:questionnaire}
\end{figure*}

\begin{figure}[]
  \centering
\includegraphics[width=\linewidth]{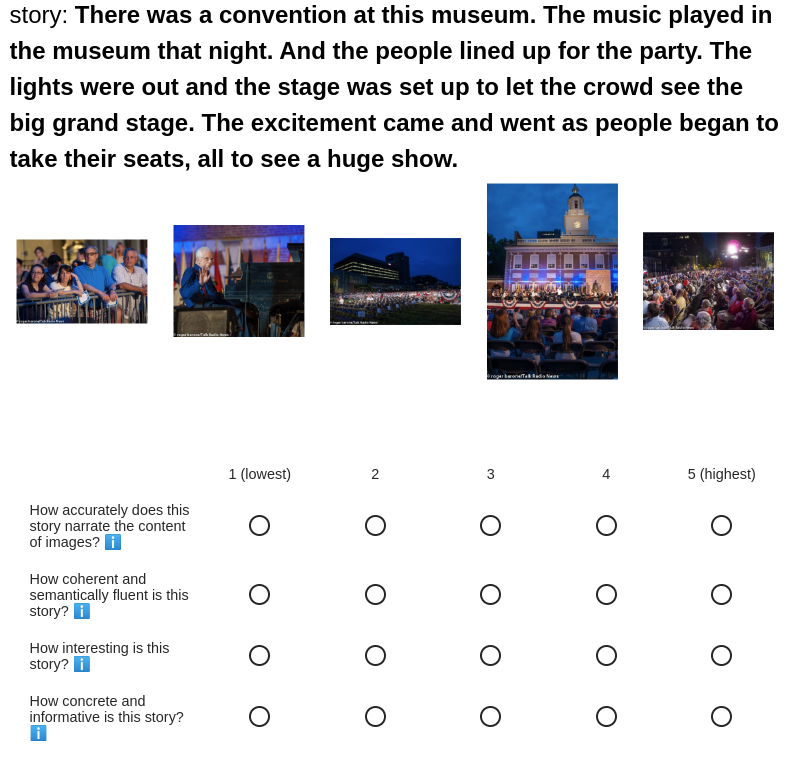}
  \caption{One example question in the human evaluation questionnaire.
  }
  \label{fig:question}
\end{figure}

\autoref{fig:questionnaire} presents the instruction, sample image sequence stories provided in the human evaluation questionnaire.
The introduction aims to make participants fully understand the specific meaning of the four evaluation criterion and the corresponding score scale. 
The samples are intended to help participants build a mental expectation of the image sequences and stories they will see, in order to avoid the order in which the images and stories appear influencing their judgment. 
In \autoref{fig:question}, we show an example question that consists of a story generated by 1 out of 8 models, a sequence of 5 images, and 4 direct rating questions.
We randomly shuffled all 100 image sequences and their corresponding 8 stories generated by different models in an even manner. In each participant's survey, which includes 32 questions, the same image sequence will not appear twice, and stories from all 8 models are included.
We only asked each participant to complete 32 questions (median completion time is 20mins 8secs), avoiding their judgment being affected due to excessively long periods of focus at a single survey task. 
We hired 75 annotators (38 females, 37 males) on Prolific at a hourly rate of £13.41, all of whom are proficient in English with at least the college education level.

\newpage

\newpage
\onecolumn
\section{Additional Results}
\label{sec:append-result}

\begin{table*}[ht]
    \centering
    \small
    \renewcommand{\arraystretch}{0.1}
    \begin{longtblr}[
  caption = {Results of our model with GPT2-xl, textual context concatenation before and after mapping network, +/- contrastive learning and +/-curriculum training.},
  label = {tab:first},
]{
  width = \linewidth,
  colspec = {Q[119]Q[62]Q[62]Q[62]Q[62]Q[62]Q[62]Q[63]Q[69]Q[83]Q[62]Q[73]Q[98]},
  row{2} = {c},
  row{7} = {c},
  row{12} = {c},
  row{17} = {c},
  column{2} = {c},
  column{3} = {c},
  column{4} = {c},
  column{5} = {c},
  column{6} = {c},
  column{7} = {c},
  column{8} = {c},
  column{9} = {c},
  column{12} = {c},
  column{13} = {c},
  cell{1}{1} = {c},
  cell{2}{1} = {c=13}{0.939\linewidth},
  cell{3}{1} = {c},
  cell{4}{1} = {c},
  cell{5}{1} = {c},
  cell{6}{1} = {c},
  cell{7}{1} = {c=13}{0.939\linewidth},
  cell{8}{1} = {c},
  cell{9}{1} = {c},
  cell{10}{1} = {c},
  cell{11}{1} = {c},
  cell{12}{1} = {c=13}{0.939\linewidth},
  cell{13}{1} = {c},
  cell{14}{1} = {c},
  cell{15}{1} = {c},
  cell{16}{1} = {c},
  cell{17}{1} = {c=13}{0.939\linewidth},
  cell{18}{1} = {c},
  cell{19}{1} = {c},
  cell{20}{1} = {c},
  cell{21}{1} = {c},
  hline{1,22} = {-}{0.08em},
  hline{2} = {1-2,4-9,12-13}{0.03em},
  hline{2} = {3,10-11}{},
  hline{3,8,13,18} = {-}{0.05em},
  hline{7,17} = {-}{},
  hline{12} = {1-9,12-13}{0.03em},
  hline{12} = {10-11}{},
}
                                                                        & B-1   & B-2   & B-3   & B-4   & M     & R-L   & CIDEr & SPICE & BLEURT & PPL   & CLIPS. & RefCLIPS. \\

+curriculum learning, +context before mapping network,-contrastive loss &       &       &       &       &       &       &       &       &        &       &        &           \\
Beam                                                                    & 62.76 & 37.95 & 22.8  & 13.91 & 32.70 & 30.53 & 12.02 & 8.49  & 31.63  & 12.23 & 63.81  & 72.24     \\
Top-$k$                                                                 & 46.34 & 20.65 & 8.22  & 3.45  & 28.17 & 21.51 & 5.83  & 7.82  & 29.09  & 32.27 & 60.73  & 69.54     \\
Nucleus                                                                 & 43.12 & 18.36 & 6.92  & 2.83  & 26.53 & 20.72 & 5.59  & 7.38  & 28.05  & 44.12 & 59.26  & 68.37     \\
SimCTG                                                                  & 56.27 & 29.84 & 14.45 & 7.07  & 27.96 & 25.98 & 8.70  & 8.87  & 30.71  & 13.39 & 62.65  & 72.35     \\

+curriculum learning, +context after mapping network,-contrastive loss  &       &       &       &       &       &       &       &       &        &       &        &           \\
Beam                                                                    & 60.19 & 35.67 & 20.45 & 13.90 & 32.52 & 27.84 & 10.95 & 8.46  & 32.37  & 11.62 & 62.63  & 72.66     \\
Top-$k$                                                                 & 52.73 & 24.91 & 10.48 & 4.67  & 26.37 & 23.05 & 4.66  & 7.51  & 29.23  & 30.22 & 61.60  & 70.13     \\
Nucleus                                                                 & 50.65 & 23.02 & 9.25  & 4.04  & 25.55 & 22.36 & 3.83  & 7.02  & 28.14  & 41.07 & 60.94  & 70.01     \\
SimCTG                                                                  & 59.76 & 32.13 & 15.43 & 7.58  & 27.13 & 25.47 & 6.94  & 8.28  & 31.19  & 12.82 & 62.88  & 72.29     \\

-curriculum learning, +context before mapping network,+contrastive loss &       &       &       &       &       &       &       &       &        &       &        &           \\
Beam                                                                    & 63.12 & 38.41 & 23.10 & 14.24 & 31.68 & 29.29 & 11.73 & 9.79  & 32.21  & 11.12 & 65.61  & 74.58     \\
Top-$k$                                                                 & 46.58 & 22.10 & 9.16  & 5.93  & 25.28 & 25.71 & 6.79  & 8.86  & 28.20  & 33.67 & 62.50  & 72.37     \\
Nucleus                                                                 & 44.91 & 20.43 & 8.19  & 4.91  & 24.26 & 23.59 & 6.27  & 8.03  & 27.13  & 40.91 & 61.89  & 71.68     \\
SimCTG                                                                  & 56.79 & 31.65 & 15.93 & 8.89  & 29.02 & 27.54 & 8.12  & 9.71  & 30.56  & 13.03 & 64.87  & 73.92     \\
-curriculum learning, +context after mapping network,+contrastive loss  &       &       &       &       &       &       &       &       &        &       &        &           \\
Beam                                                                    & 62.83 & 38.04 & 22.87 & 14.12 & 31.84 & 29.20 & 11.56 & 9.63  & 32.43  & 10.41 & 64.82  & 74.17     \\
Top-$k$                                                                 & 47.25 & 22.12 & 9.14  & 4.29  & 25.12 & 22.67 & 5.62  & 8.74  & 29.81  & 33.28 & 63.32  & 72.11     \\
Nucleus                                                                 & 44.40 & 19.76 & 7.71  & 3.73  & 24.03 & 21.75 & 4.91  & 7.72  & 28.18  & 43.92 & 62.75  & 71.04     \\
SimCTG                                                                  & 56.90 & 31.11 & 15.27 & 8.37  & 29.21 & 26.32 & 7.88  & 9.65  & 31.08  & 12.46 & 64.59  & 73.72     
\end{longtblr}
\end{table*}

\newpage
\begin{table*}[ht]
\small
\centering
 \setlength\extrarowheight{-3pt}
\begin{longtblr}[
  caption = {Results of our model with different GPT2 language models, textual context concatenation after mapping network, and without contrastive learning and curriculum training.},
  label = {tab:first},
]{
  width = \linewidth,
  colspec = {Q[119]Q[62]Q[62]Q[62]Q[62]Q[62]Q[62]Q[63]Q[69]Q[83]Q[62]Q[73]Q[98]},
  row{2} = {c},
  row{7} = {c},
  row{12} = {c},
  row{17} = {c},
  column{2} = {c},
  column{3} = {c},
  column{4} = {c},
  column{5} = {c},
  column{6} = {c},
  column{7} = {c},
  column{8} = {c},
  column{9} = {c},
  column{12} = {c},
  column{13} = {c},
  cell{1}{1} = {c},
  cell{2}{1} = {c=13}{0.939\linewidth},
  cell{3}{1} = {c},
  cell{4}{1} = {c},
  cell{5}{1} = {c},
  cell{6}{1} = {c},
  cell{7}{1} = {c=13}{0.939\linewidth},
  cell{8}{1} = {c},
  cell{9}{1} = {c},
  cell{10}{1} = {c},
  cell{11}{1} = {c},
  cell{12}{1} = {c=13}{0.939\linewidth},
  cell{13}{1} = {c},
  cell{14}{1} = {c},
  cell{15}{1} = {c},
  cell{16}{1} = {c},
  cell{17}{1} = {c=13}{0.939\linewidth},
  cell{18}{1} = {c},
  cell{19}{1} = {c},
  cell{20}{1} = {c},
  cell{21}{1} = {c},
  hline{1,22} = {-}{0.08em},
  hline{2} = {1-2,4-9,12-13}{0.03em},
  hline{2} = {3,10-11}{},
  hline{3,8,13,18} = {-}{0.05em},
  hline{7,17} = {-}{},
  hline{12} = {1-9,12-13}{0.03em},
  hline{12} = {10-11}{},
}
                     & B-1   & B-2   & B-3   & B-4   & M     & R-L   & CIDEr & SPICE & BLEURT & PPL   & CLIPS. & RefCLIPS. \\
GPT2-\textbf{small}  &       &       &       &       &       &       &       &       &        &       &        &           \\
Beam                 & 23.63 & 10.53 & 5.26  & 3.00  & 7.16  & 10.47 & 11.41 & 4.66  & 26.46  & 13.90 & 53.52  & 60.73     \\
Top-$k$              & 24.75 & 13.73 & 5.88  & 3.78  & 9.89  & 17.97 & 6.62  & 4.96  & 24.07  & 43.87 & 50.87  & 59.26     \\
Nucleus              & 26.44 & 13.90 & 5.72  & 4.13  & 10.05 & 16.85 & 5.98  & 5.14  & 28.20  & 53.99 & 50.74  & 58.96     \\
SimCTG                & 26.92 & 14.19 & 6.05  & 4.38  & 10.76 & 16.92 & 5.48  & 4.91  & 25.59  & 22.53 & 51.18  & 59.44     \\
GPT2-\textbf{medium} &       &       &       &       &       &       &       &       &        &       &        &           \\
Beam                 & 33.16 & 15.80 & 8.45  & 4.77  & 9.88  & 22.79 & 18.37 & 7.22  & 28.63  & 13.25 & 57.30  & 63.48     \\
Top-$k$              & 31.83 & 13.86 & 6.58  & 3.29  & 9.24  & 22.25 & 6.91  & 6.74  & 26.09  & 40.23 & 56.47  & 64.12     \\
Nucleus              & 30.49 & 13.58 & 5.87  & 3.45  & 8.93  & 21.18 & 6.33  & 6.05  & 25.05  & 56.75 & 55.85  & 63.91     \\
SimCTG                & 34.81 & 16.76 & 7.58  & 4.18  & 9.42  & 23.35 & 12.90 & 7.63  & 28.71  & 21.59 & 57.19  & 63.93     \\
GPT2-\textbf{large}  &       &       &       &       &       &       &       &       &        &       &        &           \\
Beam                 & 56.67 & 33.23 & 19.48 & 11.50 & 13.36 & 28.71 & 18.40 & 7.66  & 31.19  & 11.36 & 61.22  & 71.15     \\
Top-$k$              & 51.64 & 24.50 & 10.45 & 4.51  & 13.68 & 24.23 & 8.41  & 7.72  & 28.17  & 35.11 & 59.54  & 69.35     \\
Nucleus              & 49.71 & 22.76 & 9.41  & 4.27  & 13.14 & 23.41 & 6.37  & 7.12  & 27.07  & 50.06 & 58.37  & 68.19     \\
SimCTG                & 59.08 & 32.34 & 15.99 & 7.95  & 13.82 & 27.41 & 12.59 & 7.98  & 30.64  & 19.62 & 61.34  & 71.14     \\
GPT2-\textbf{xl}     &       &       &       &       &       &       &       &       &        &       &        &           \\
Beam                 & 62.88 & 38.04 & 22.96 & 14.01 & 14.95 & 29.30 & 17.64 & 9.37  & 32.37  & 10.73 & 62.08  & 71.77     \\
Top-$k$              & 55.76 & 28.01 & 12.74 & 5.89  & 13.13 & 25.67 & 5.61  & 8.61  & 29.23  & 35.68 & 60.06  & 69.75     \\
Nucleus              & 49.29 & 22.55 & 9.88  & 4.93  & 12.86 & 23.60 & 3.86  & 7.36  & 28.14  & 46.17 & 59.16  & 68.81     \\
SimCTG                & 60.52 & 33.76 & 17.19 & 8.92  & 13.65 & 27.48 & 8.01  & 9.18  & 31.09  & 13.92 & 62.02  & 71.66     
\end{longtblr}
\end{table*}

\newpage
\section{Human Evaluation Significance Test}
\label{sec:append-tukey}
\begin{figure*}[!htb]
\minipage{0.49\textwidth}
  \includegraphics[width=\linewidth]{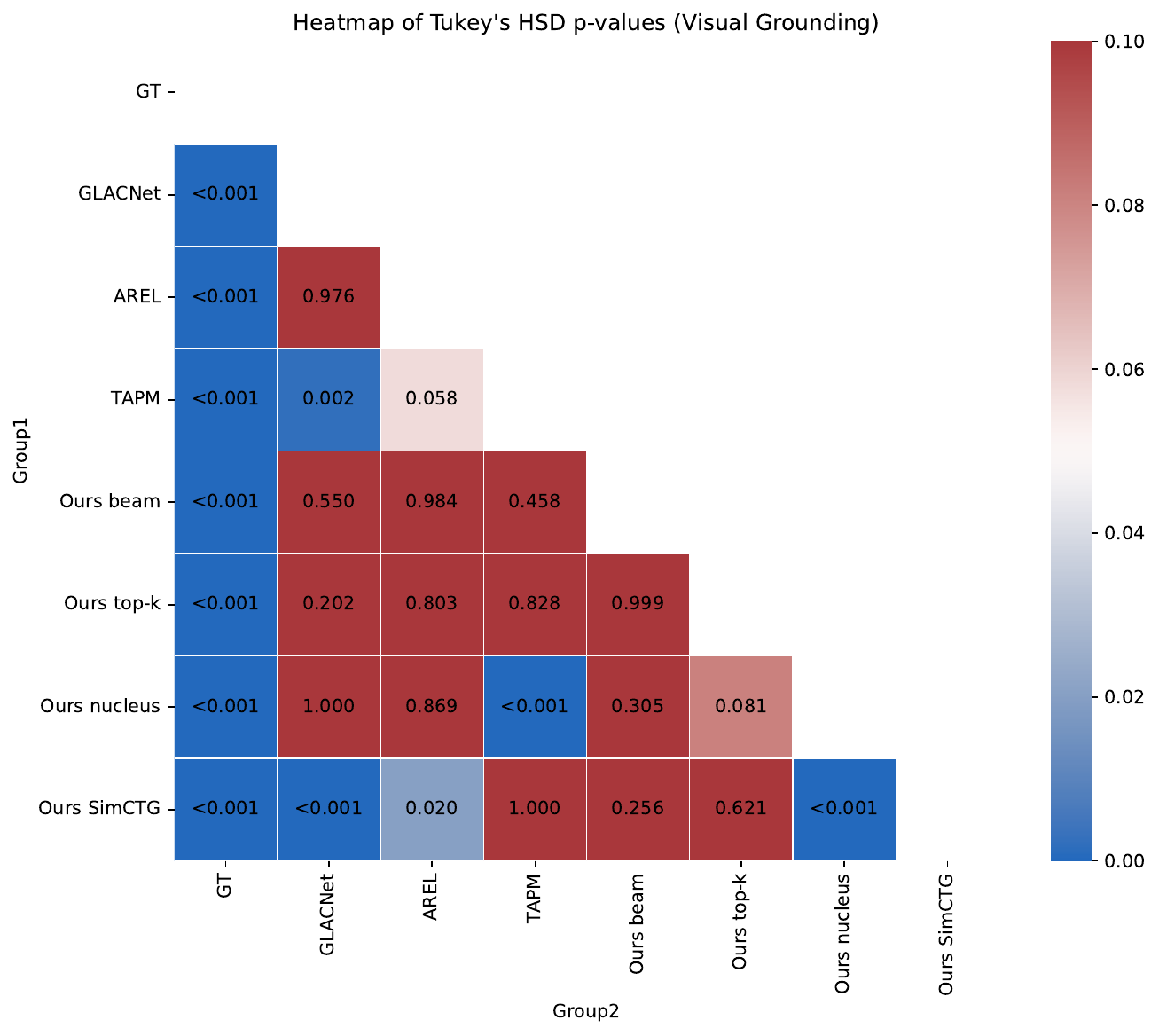}
  \caption{Visual Grounding}
\endminipage\hfill
\minipage{0.49\textwidth}
  \includegraphics[width=\linewidth]{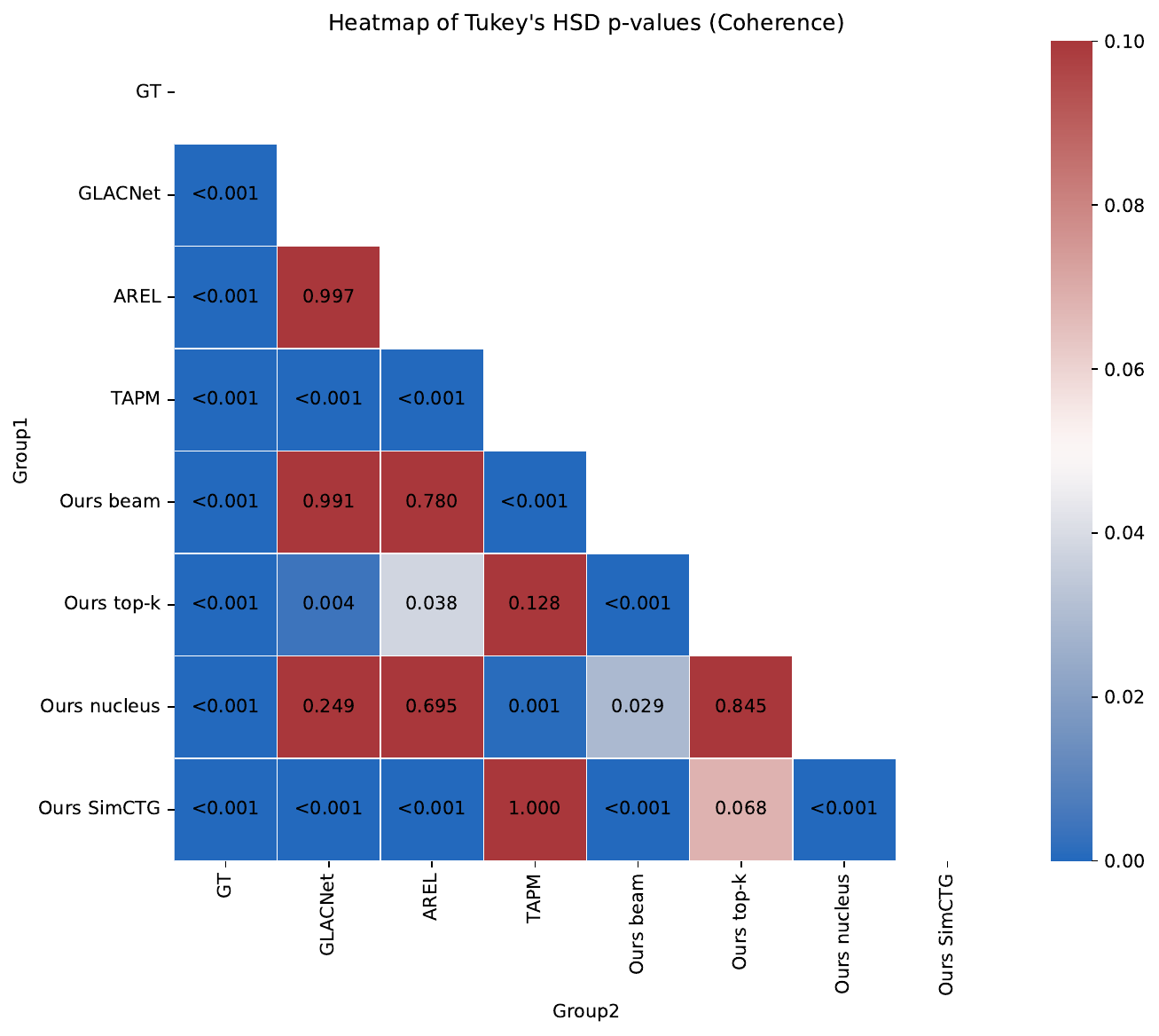}
  \caption{Coherence}
\endminipage\hfill
\minipage{0.49\textwidth}
  \includegraphics[width=\linewidth]{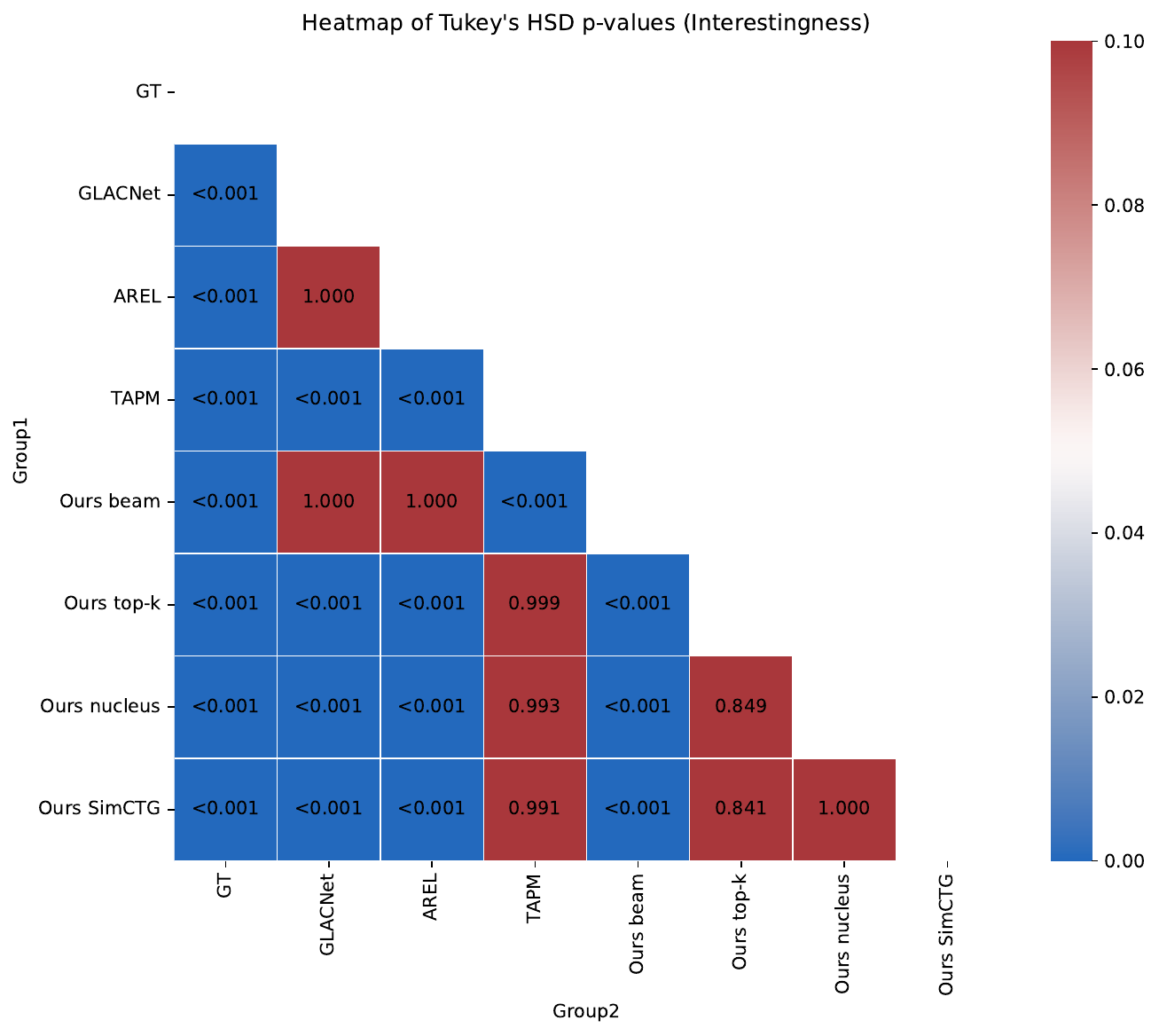}
  \caption{Interestingness}
\endminipage\hfill
\minipage{0.49\textwidth}
  \includegraphics[width=\linewidth]{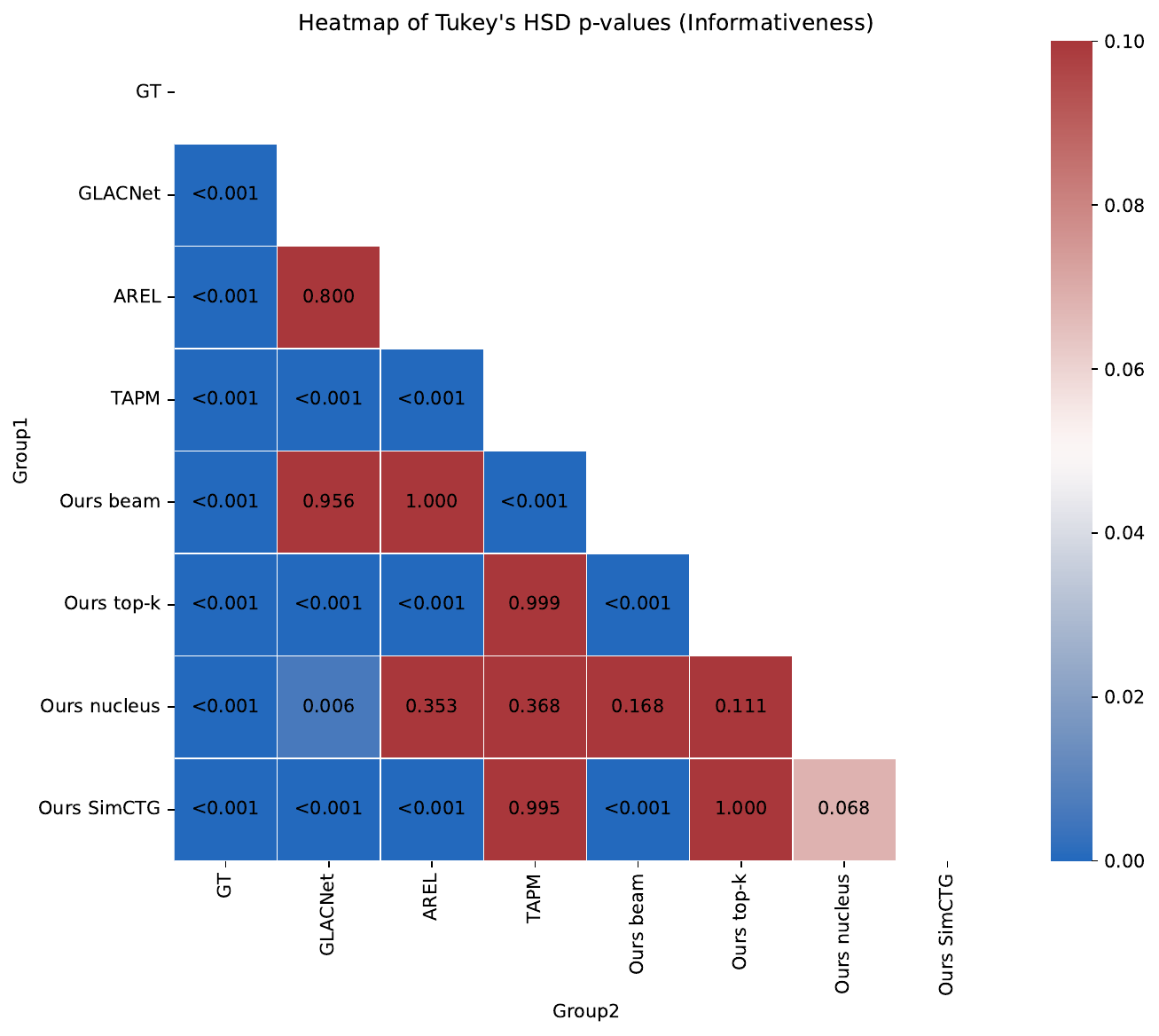}
  \caption{Informativeness}
\endminipage
  \caption{$p$-values of Tukey's HSD Pairwise Group Comparisons (95.0\% Confidence Interval)}
  \label{fig:tukey-sig}
\end{figure*}

We conduct Tukey's HSD pairwise group comparisons of human evaluation scores we collected as shown in \autoref{fig:tukey-sig}.

\end{document}